# Using Deep Learning to Detecting Deepfakes

Jacob Mallet
*Department of Computer Science*
*University of Wisconsin – Eau Claire*
Eau Claire, WI, USA
malletjc3227@uwec.edu

Dr. Rushit Dave
*Department of Computer Information Science*
*Minnesota State University, Mankato*
Mankato, MN, USA
rushit.dave@mnsu.edu

Dr. Naeem Seliya
*Department of Computer Science*
*University of Wisconsin – Eau Claire*
Eau Claire, WI, USA
seliyana@uwec.edu

Dr. Mounika Vanamala
*Department of Computer Science*
*University of Wisconsin – Eau Claire*
Eau Claire, WI, USA
vanamalam@uwec.edu

*Abstract*—In the recent years, social media has grown to become a major source of information for many online users. This has given rise to the spread of misinformation through deepfakes. Deepfakes are videos or images that replace one person's face with another computer-generated face, often a more recognizable person in society. With the recent advances in technology, a person with little technological experience can generate these videos. This enables them to mimic a power figure in society, such as a president or celebrity, creating the potential danger of spreading misinformation and other nefarious uses of deepfakes. To combat this online threat, researchers have developed models that are designed to detect deepfakes. This study looks at various deepfake detection models that use deep learning algorithms to combat this looming threat. This survey focuses on providing a comprehensive overview of the current state of deepfake detection models and the unique approaches many researchers take to solving this problem. The benefits, limitations, and suggestions for future work will be thoroughly discussed throughout this paper.

## I. Introduction

Over the past few years, technology has advanced in a way that enables users with little technological background to effortlessly create a deepfake. A deepfake is a video or image that replaces one person's face with a synthetic, computer-generated version of the authentic person. Deepfakes, commonly created by a generative adversarial network (GAN), hold the capability of fooling an audience into believing a recognizable figure in society is speaking, when they are not. As society continues to heighten its dependency on social media for providing important information, the threat of misinformation stemming from deepfakes continues to grow. People like presidents, celebrities, and other powerful figures in society can all have their likeness mimicked in a deepfake video that is often indistinguishable from the original to the human eye, creating the potential of misuse. This imminent threat of misinformation has produced the need for software to be developed that could uncover these deepfakes that the human eye is unable to detect. As artificial intelligence continues to advance in related fields, as seen in [1-9], it's clear this is the tool necessary to combat deepfake misuse.

The main idea behind detecting a deepfake is to find an inconsistency between images from the person of interest (POI) and the face generated by the GAN, but researchers have still taken a wide variety of approaches to this issue. Most deepfake models that attempt to distinguish between a deepfake and an authentic video can fall into three categories, which all take advantage of deep learning in some way. The first type of methodology can be seen in [10-13], as some researchers have attempted to find the inconsistencies of how the POI's facial features, and other regions of interest move compared to the deepfake's constructed movement of the same features across the temporal domain. These models essentially analyze the movement of facial features between frames and attempt to identify movement that is inconsistent with the authentic POI. Another way of approaching the issue, like [14, 15] do, is to strive to detect deepfakes by searching for biological signs of authentic human life in a face. An example of this, specifically in [15], can be trying to identify color changes in the skin to conclude if there is blood under the tissues of the face, like there would be in an authentic face. A different approach taken by [16] uses the audio portion, as well as other features, of a deepfake to distinguish the authentic POI video from a deepfake. A major problem that plagues deepfake detection models is their transferability, highlighted by [17]. Many are unable to perform at the same level when applied in the real world or different datasets, but [18] looks to solve this issue by inputting a wide variety of features. All in all, there are numerous ways to approach the issue of detecting deepfakes generated by a GAN. In this work, we will provide an overview of the current state of deepfake detection models, as well as providing several different approaches to this topic.

## II. Background

### A. Database

A necessary tool in creating a Deepfake detection model is the existence of novel datasets that can train and test models. Using a GAN, researchers in [19] do just this and create a database of Deepfakes with the sole purpose to aid others in developing detection models. Future research provided by [20], created another dataset of Deepfakes titled "Celeb-DF v2". [20] improved upon existing datasets later by improving the methodology in creating these Deepfakes, that increases the difficulty of detecting these Deepfakes by using an improved synthesis algorithm. [21] develops another widely used database used to benchmark researcher's models, as the database stores 1.8 million manipulated images. Creating an extensive database of Deepfakes that can be used as a universal benchmark for models is essential to advance research in this field.

### B. Deep Learning Networks

One similarity found in many of the recent deep learning Deepfake detection models, such as [22-27], comes from the use of a convolutional neural network (CNN) in some way throughout the entirety of the model's architecture. A CNN is a deep learning classification neural network that contains multiple different layers, where each one provides an additional element to the CNN, with the end goal being able to identify portions of an image that are critical in achieving an accurate classification. The first type of layer, the Convolutional layer, is responsible for extracting critical portions of the frames. A matrix, referred to as the Kernel, traverses through the image, jumping a specified distance, known as the stride, performing matrix multiplication operations on each portion of the image it visits. Essentially, in the convolutional layer, the CNN can take the inputted frame and identify the high-level regions of dependence. The second type of a layer is called the Pooling layer, which is responsible for altering the size of the features extracted. Similar to the Convolutional layer, this layer also traverses the image, only this time the size of the output provided is less complex. This does come at a cost, as information is lost in exchange for more simple and efficient output, however there are benefits to this layer as well. The Pooling layer reduces the complexity and dimensionality, making the model more efficient. Multiple layers of these two can be stacked upon each other to dive deeper into uncovering smaller features as well. The final layer, called the Fully-Connected layer, performs the final steps of the network. Using a softmax activation function, the information gathered from the previous layers is used to make a classification decision. CNNs are becoming increasingly common due to their success in identifying and extracting regions of an image that are important in classification.

Deepfake models commonly take advantage of CNNs ability to identify areas of the frame that have high spatial and temporal dependencies. This may result in the extraction of a face from an image, or various features of a face, such as the eyes or mouth of a person, which aids in identifying potential inconsistencies between the authentic image and the deepfake. While a large portion of Deepfake models may utilize a CNN, as seen in [22-27], they are applied in unique ways for each model.

### III. LITERATURE REVIEW

### A. Temporal Feature Deepfake Detection Methods

Researchers in [28] aim to distinguish between deepfakes and authentic images by searching for inconsistencies among the deepfake from frame to frame. First, a Multitask Cascaded Convolutional Neural Network is used to extract just the face from the frame. After the frame only contains a face, a sequence of 16 and 32 frames are fed to a Long Short-Term Memory (LSTM) network to capture information across the temporal domain. Vectors of information are passed from the previous step to the next to sufficiently analyze the sequence of movements from the video. A 3D convolutional network is then applied to uncover any poses that have not been identified from the authentic source video. Where [28] really separates itself from other work is by using a triplet network. This network groups features together and performs various functions on the distance between them to classify the deepfake.

Another unique approach to detecting deepfakes comes from [29]. Researchers construct a model that involves three different CNN streams, with each stream receiving different input. The first stream is given the original image, and it's responsible for learning general characteristics of the person, which includes head shape and hair color. This stream is built on a pre-trained VGG16 model, which was developed in [30]. The second stream is given a blurred version of the image and must identify the skin color. Similar to the previous stream, this one is built on a pre-trained VGG19 model, which was also developed in [30]. The last CNN stream is fed a sharpened version of the image, and deals with local features of the face. This stream is built on a pre-trained ResNet18 model, which is another deep CNN model that is composed of 71 layers. [29] sharpens the image by increasing the contrast around the areas where colors intersect. These three streams are combined in the fusion layer at the feature level, and a decision is made in the classification output layer.

A methodology involving analyzing the motion between subsequent frames is applied in the model developed by [31]. [31] hypothesizes that when an attacker creates a deepfake, they must decode the original video, then place it back into a compressed video format. The goal of this paper is to exploit temporal correlation between consecutive frames. Inter-frame prediction is carried out to determine a prediction error, which will be fed to deep learning algorithms. 47 features are computed from the prediction error, including mean, variance, skewness, energy, and more. Two approaches from here are experimented with, the first being a spatial-based CNN approach. Two different CNNs are set up in the first approach, with one receiving a bi-dimensional matrix composed of the prediction error, and the other getting a vector of the 47 elements computed from the prediction error. The second approach in [31] employs a LSTM in a more sequence-based approach. Their goal for this methodology is to employ a LSTM model that is capable of uncovering long-term dependencies throughout an entire video sequence, rather than strictly analyzing consecutive frames, which is a more short-term period. The input for this model is similar to the last approach

Funding provided by University of Wisconsin-Eau Claire's Office of Research and Supported Projects

where a prediction error matrix is given to the LSTM. For each input sequence, the probability of it being altered or original is generated using a sigmoid activation function for their output layer. An additional LSTM developed for this experiment is briefly reviewed in [31], where the input given here is the mono-dimensional features at each time t, with a fully connected layer to generate binary prediction outcomes.

## B. Biological Feature Deepfake Detection Models

One different way to detect a deepfake is by identifying biological traits only an actual human would possess. [14] attempts to do this by locating signals of human life in a video that aren't visible to the human eye. This can include anatomical actions such as a heartbeat, breathing, or blood flow. Such signals can be found by deep learning algorithms, as the slightest change of light or reflection in the face can be key in detecting a deepfake. Their model will extract photoplethysmography (PPG) cells, which are where these signals can be found. However, not all PPG cells are created equal, so extracting regions of interest that have the highest stability of PPG signals possible is key. [14] identifies this target region to be located between the eye and mouth regions, which maximizes the skin exposure. These nonlinear regions of interest were then aligned into a rectangular image, which then are divided into 32 equal squares to calculate the raw "Chrom-PPG", and finally information from the frequency domain is added into these cells. A CNN using 4 VGG19 blocks from [30] are used for the implementation of this methodology to classify the data.

Similar to the previous model, [15] attempts to distinguish between genuine and deepfake faces by analyzing the face for signs of human life. This study searches for information relating to the heart rate of a person by using remote photoplethysmography (rPPG). rPPG methods are used to uncover minute changes in the color of human skin, which would conclude that there is human blood underneath the tissues of the face. The model is built on a Convolutional Attention Network (CAN) that is composed of two different CNN branches. One branch, named the Appearance Model, focuses on the individual frames themselves, attempting to identify key areas of interest, which it will supply to the other branch. The other branch, named the Motion Model, is responsible for detecting changes between subsequent frames from the video. Their classification layer, also the final layer of the model, uses a sigmoid activation function to obtain a score ranging from 0 to 1. This score represents the probability that the face being analyzed is real.

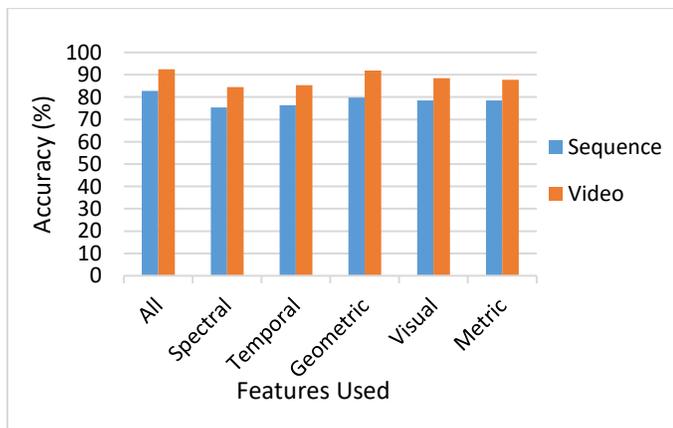

Fig. 1. Results from Combining Different Features [31]

TABLE I. COMPARINNG THE AREA UNDER CURVE (AUC)

| References | Models Evaluated | |
|---|---|---|
| | *Dataset* | *AUC* |
| [15] | Celeb-DF v2 | 99.9% |
| [22] | DeepfakeTIMIT | 99.9% |
| [33] | FaceForensics++ | 99.5% |
| [33] | Celeb-DF | 83.6% |
| [34] | FaceForensics++ | 100% |

## IV. DISCUSSION & ANALYSIS

### A. Databases

In order to properly compare models to each other, evaluating them on the same, or at least similar databases is a key in doing so. Researchers in [19] create a large, robust dataset available for existing or future research involving deepfake detection models to utilize in their work. The goal of Celeb-DF was to improve upon the existing deepfake synthesis algorithms out there in several ways. Some deepfake detection models, [35], strictly search for a footprint that the deepfake generation left behind, so constantly improving the generation method and eliminating these signals left behind is key, like researchers in [36] also do. As most deepfakes are 64 x 64 or 128 x 128 pixels, [19] aims to synthesize the deepfakes to be 256 x 256 pixels to improve the resolution. This not only increases the quality but reduces the impact of resizing and rotation operations throughout the video. The next improvement carried out involves matching and blending colors to match the skin tone of the person the deepfake is being applied to more accurately than previous generation methods. Authors in [19] do this by applying a color transfer algorithm from the authentic face to the generated deepfake, which minimizes color mismatch on the face. [19] also identifies the size of the synthesized face mask to be a common issue among deepfakes. This problem is dealt with by making the mask too large to begin with, then identifying facial landmarks to trim the size of the mask down to, as previous generation methods have left too much of the person's face showing where the deepfake should be. The final improvement in [19] removes any temporal flickering of the facemask during periods of movement or rotation. Temporal landmarks among the synthetic face are identified and the correlations throughout the video are kept the same. [19] uses a metric know as the Mask-SSIM score, which is referenced as a quantitative metric that evaluates the visual quality of a deepfake's video frames. Compared to the other datasets evaluated in the study, Celeb-DF, scored the highest with an average score of 0.92. All in all, researchers in [19] created a novel dataset consisting of 5639 deepfake videos, along with 590 real videos, available for researchers to evaluate their own models and compare them against others using this shared dataset as a benchmark.

TABLE II.    RESULTS USING CELEB-DF DATASET

| References | Models Evaluated | |
| --- | --- | --- |
| | *Methodology* | *Accuracy* |
| [14] | Biological Features | 92.17% |
| [15] | Biological Features | 98.7% |
| [28] | Temporal | 96% |
| [29] | Temporal | 99.95% |
| [32] | Biological Features & Temporal | 88.35% |
| [37] | Temporal | 94.21% |

*B. Results*

In this survey, a variety of deep learning models used to detect deepfakes were introduced and evaluated. These models were tested on commonly used datasets named Celeb DF and FaceForensics++, which provides a good benchmark to compare models to each other, as results can be heavily dependent on the dataset applied. Looking at the results from a high level, models using temporal features produced the highest accuracies for both datasets observed, although only a few models that primarily used biological features were present in this survey. While accuracies were generally high across the entirety of the models reviewed, [15], [29], and [33] really stood out from the others. These models all achieved an accuracy of at least 98.7%, with [29] yielding the highest accuracy of any model, at 99.95%. Another interesting take away from the results shown in Table 2 and Table 3, is the lowest performance in the results from [32], which uniquely utilized biological and temporal features. In Table I, AUC scores were very high for most of the models observed, as 4 out of 5 of the models scored at least 99.5%. These results display the effectiveness of CNNs and other deep learning algorithms. All in all, the models scored high overall and displayed the promise of the existing tools to combat deepfakes on the internet.

TABLE III.    RESULTS USING FACEFORENSICS++ DATASET

| References | Models Evaluated | |
| --- | --- | --- |
| | *Features Used* | *Accuracy* |
| [14] | Biological Features | 93.69% |
| [28] | Temporal | 86.74% |
| [31] | Temporal | 94.29% |
| [32] | Biological Features & Temporal | 92.48% |
| [34] | Temporal | 99.65% |
| [38] | Temporal | 84% |
| [39] | Temporal | 81.61% |

V. LIMITATIONS & CONCLUSION

Detecting deepfakes has become a significant problem in the online world, yet it remains one of the more challenging issues to solve. As social media continues to become a mass information source for many, the threat of deepfakes being used nefariously grows, as well as the potential consequences. Some detection tools are available to the public already, overviewed in [40], yet there is still room for much improvement. One issue that plagues many deepfake models is their lack of transferability. Many deepfake generation methods leave their own unique traces behind and can be learned by a deep learning algorithm, but applying that same model to deepfakes generated differently can cause accuracies to drop. Another problem that may be more relevant in future, real world application scenarios, is the cost of many of these models. In order to be able to eventually aid the public in detecting deepfakes spreading misinformation, the models must be able to be run by a standard individual on a user's device. Finally, one limitation that caused issues in this study and not the models themselves, is how frequently papers use differing success metrics, which increases the difficulty in comparing models to each other.

Detecting deepfakes remains a key issue to solve with society's rapidly growing dependence on social media. The potential of nefarious misuse of deepfakes is immense, and it only grows as generating deepfakes become more convenient and accessible for even those with limited technological knowledge. Future work should include developing models that are able to perform well on multiple different databases in order to prepare this technology to be applied in a real-world context and aid people online around the world.

ACKNOWLEDGMENT

Funding for this project has been provided by the University of Wisconsin-Eau Claire's Office of Research and Special Programs Summer Research Experience Grant